\documentclass{article}
\usepackage{spconf,amsmath,graphicx}

\usepackage{times}
\usepackage{epsfig}
\usepackage{graphicx}
\usepackage{amsmath}
\usepackage{amssymb}
\usepackage{booktabs}

\usepackage{subcaption} 

\usepackage{amsfonts}

\usepackage{sansmath}

\usepackage{algorithm}
\usepackage{algpseudocode}

\makeatletter
\renewcommand{\ALG@name}{Recipe}
\makeatother

\usepackage[dvipsnames]{xcolor}

\usepackage{amsfonts}
\usepackage{amsthm}

\usepackage{tikz-cd}
\usepackage{mathtools}
\usepackage{enumitem}
\usepackage{faktor}
\usepackage{tabularx}
\usepackage{hyperref}
\usepackage{placeins}

\usepackage{float}

\usepackage{tikz}
\usepackage{comment}
\usepackage{amsmath,amssymb} 
\usepackage{color}
\usepackage{url}
\usepackage{dsfont}
\usepackage{soul}

\usepackage{afterpage}

\usepackage{tabularx}

\definecolor{palepurple}{RGB}{220, 198, 224}
\definecolor{paleyellow}{RGB}{255, 255, 153}
\definecolor{orange}{RGB}{255, 153, 0}
\definecolor{purple}{rgb}{0.5, 0.0, 0.5}

\usepackage{bbding}

\usepackage{relsize}
\newcommand*{\defeq}{\stackrel{\mathsmaller{\mathsf{def}}}{=}}

\definecolor{applegreen}{rgb}{0.55, 0.71, 0.0}
\definecolor{babyblueeyes}{rgb}{0.33, 0.49, 0.65}
\definecolor{candypink}{rgb}{0.89, 0.44, 0.48}

\usepackage{amsmath}

\newcommand{\RADIN}{$\mathsf{RADIN}$}

\usepackage{amsmath}
\usepackage{amsthm}

\definecolor{pink}{rgb}{1.0, 0.75, 0.8}
\definecolor{piggypink}{rgb}{0.99, 0.87, 0.9}
\usepackage{soul} 
\sethlcolor{pink}

\usepackage{ifsym}

\usepackage{bbold}
\usepackage{dsfont}

	\DeclareMathOperator{\ind}{ind}

	\DeclareMathOperator{\Hom}{Hom}
	
	\DeclareMathOperator{\init}{init}

	\DeclareMathOperator{\ens}{ens}
	\DeclareMathOperator{\soup}{soup}

\title{\underline{\RADIN}: Souping on a Budget}

\name{
    \parbox{\linewidth}{\centering
    ~~~~~~~~~Thibaut Ménès$^{1\ast}$ \qquad
    Olivier Risser-Maroix$^{2\ast \dag}$ 
    }
   \thanks{$\ast$ Equal contribution.\\ \indent~~~$\dag$ Corresponding author.}
}

\address{
    \vspace{1mm}$^1$Université Sorbonne Paris Nord 
	Laboratoire Analyse, Géométrie et Applications, France\\
    \vspace{0.5mm}$^2$Artinity, Paris, France\\
    \vspace{0.5mm}
    \normalsize\texttt{menes@math.univ-paris13.fr, orissermaroix@gmail.com}
}

\begin{document}
%
\maketitle
\begin{abstract}
Model Soups, extending Stochastic Weights Averaging (SWA), combine models fine-tuned with different hyperparameters. Yet, their adoption is hindered by computational challenges due to subset selection issues. In this paper, we propose to speed up model soups by approximating soups performance using averaged ensemble logits performances. 
Theoretical insights validate the congruence between ensemble logits and weight averaging soups across any mixing ratios 
Our \textbf{R}esource \textbf{AD}justed soups craft\textbf{IN}g (\RADIN) procedure stands out by allowing flexible evaluation budgets, enabling users to adjust his budget of exploration adapted to his resources while increasing performance at lower budget compared to previous greedy approach (up to 4\% on ImageNet). 
\end{abstract}

\begin{figure*}[htb]
\begin{minipage}[b]{.5\linewidth}
  \centering
  \centerline{\includegraphics[height=5.4cm]{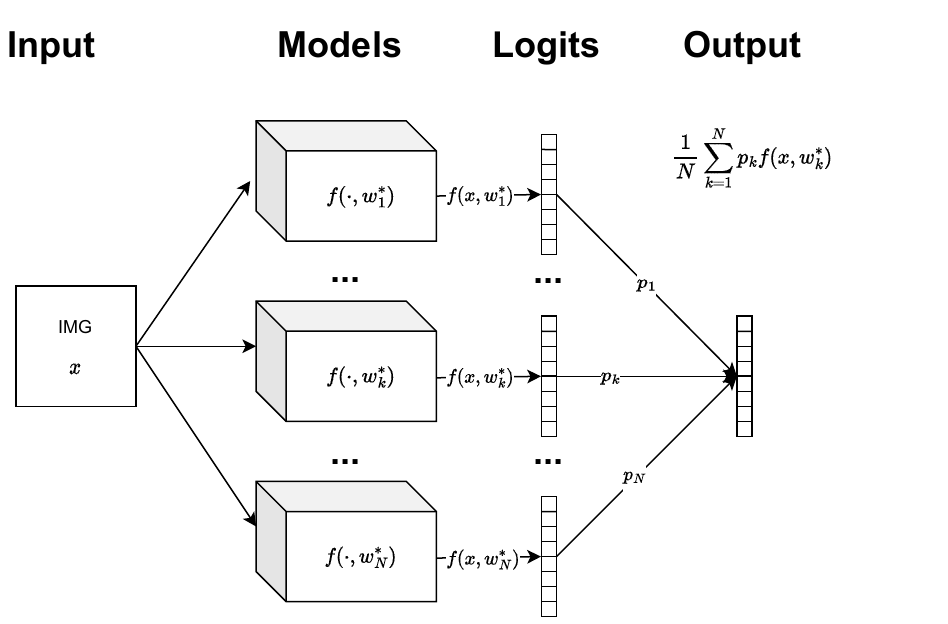}}
  \centerline{(a) Ensemble}\medskip
\end{minipage}
\hfill
\begin{minipage}[b]{0.5\linewidth}
  \centering
  \centerline{\includegraphics[height=5.6cm]{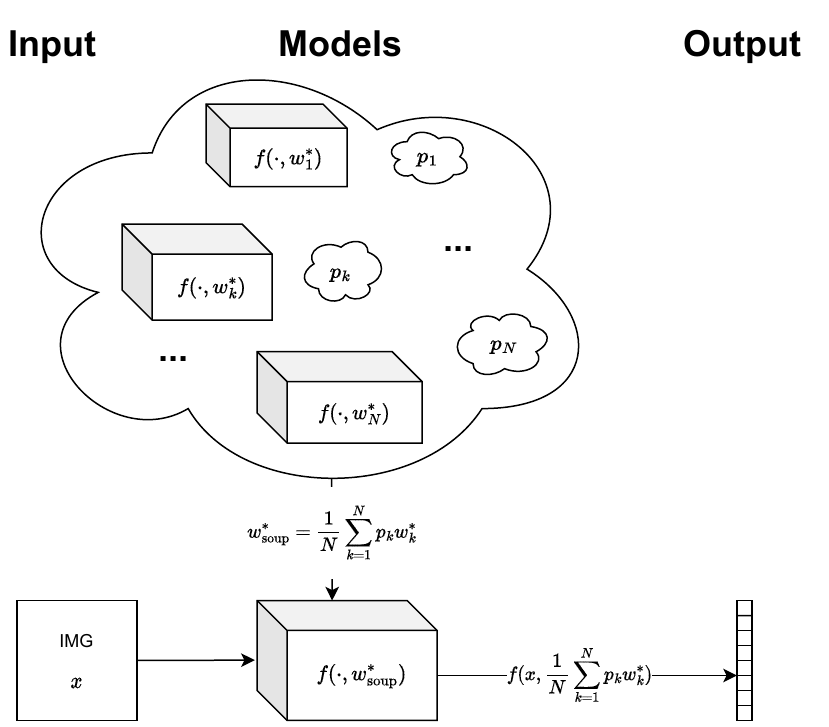}}
  \centerline{(b) Soup}\medskip
\end{minipage}
\caption{Difference between model ensembling at logit level (a) and model souping (b). While the logits ensembling requires $O(N)$ full models inferences for a single image before averaging predicted logits, the soup alternative proposes to approximate the ensemble outputs in $O(1)$ by averaging before the $N$ weights of the ensemble into a single model.}
\label{fig:res}
\end{figure*}

\section{Introduction}
\label{sec:intro}

Deep learning conventionally involves training a singular neural network across multiple epochs, selecting the top-performing iteration based on validation scores for deployment. 
However, this approach is not without its challenges. 
Notably, individual models can be outperformed by ensembles—even those comprised of weaker models. 
Furthermore, models trained via classical Empirical Risk Minimization (ERM) often lack robustness to distribution shifts \cite{shah2020pitfalls}. 

Ensembling, an entrenched technique in machine learning, aims to boost performance and reduce variance by aggregating predictions from multiple models. 
Random Forest stands out as a notable representative of this methodology \cite{Breiman2001RF}. 
Yet, for an ensemble of \(N\) models, the inference time, scaling as \(O(N)\), remains a significant cost.

An alternative technique, aiming to emulate ensemble performance with the computational simplicity of single-model inference, \(O(1)\), leverages the averaging of fine-tuned weights from models sharing identical architecture and initial pre-training. 
This technique, known as Stochastic Weights Averaging (SWA) \cite{Izmailov2018AveragingWL}, has been further evolved into \textit{model soups} \cite{wortsman2022model}. 
Model soups have exhibited superior performance compared to the top-performing individual models from an ensemble. Their construction involves selecting a subset of the \(N\) models to average. 
However, identifying the optimal subset theoretically necessitates \(2^N\) evaluations, with each evaluation being costly: a full pass over the validation set. 
Consequently, the original authors introduced a heuristic $-$ a greedy algorithm $-$ to approximate the best subset using a fixed budget of \(N\) operations. 

Our contribution lies in extending soup crafting to any budget. 
We present an adaptable procedure to craft soups based on a predefined budget \(B\): low budget for faster soup crafting, higher budget for more performing soup.  
Our approach approximates the full inference evaluation of a soup on a validation set via a weighted average of ensemble logits, derived from an identical model subset as the given soup. 
We provide a theoretical foundation for our method, demonstrating the performance equivalence $-$ up to the first order of a Taylor expansion $-$ between any soup and any ensemble drawn from the same subset of \(N\) models. 
We corroborate our claims with experimental results.

\section{Related Work}
\label{sec:related}

It has long time been observed that ensembles of models improve predictive performance \cite{ensemble2000, Lee2015WhyMH}. 
However, ensembles of models suffer from high inference costs by being proportional to the number of models. 

Several attempts in increasing performance of a single model by regularization have been proposed. 
One of them is \textit{dropout} \cite{srivastava14dropout}. 
Interestingly, dropout can be viewed as a way of emulating ensembles at lower cost, since only a single model is required. 

Earlier work introduced weights averaging \cite{Ruppert1988EfficientEF, Polyak1992AccelerationOS, utans1996weight}, decades ago. 
More recently, approaches based on weights averaging started to re-emerge \cite{Izmailov2018AveragingWL, wortsman2022model}. 
A simple average of models weights fine-tuned from a same pre-trained point is supported by the fact that model averaging is asymptotically better than model selection for prediction \cite{Le2022ModelAI}. 
Indeed, \cite{Mandt2017StochasticGD} has shown, under simplifying assumptions, that SGD with a fixed learning rate approximately samples from a Gaussian distribution centered at the minimum of the loss.

Exploiting findind from \cite{Mandt2017StochasticGD}, a succesfull attempt of weights averaging is proposed by \cite{Izmailov2018AveragingWL}, often referred as Stochastic Weights Averaging (SWA). 
Authors shown that simple averaging of multiple points along the trajectory of SGD, with a cyclical or constant learning rate, leads to better generalization than conventional training.

Close to our work, in the continuity of SWA \cite{Izmailov2018AveragingWL}, \cite{li2023trainable} proposes to average historical solution weights of a training procedure in a trainable fashion, rather than using fixed averaging (e.g., SWA) or a heuristic combination (e.g., Exponential Moving Average - EMA).
The purpose being to speed up the training and meanwhile improving the performance, by utilizing  early solutions in DNNs’ training. 
We differ in that their contribution focus on combining historical solutions from a single configuration while we focus on ensemble of models fine-tuned separately with different configuration of hyperparameter (or even auxiliary tasks) and thus extend the more general case of model soups and model ratatouilles.

This idea has been extended to \textit{model soups} \cite{wortsman2022model} where authors found that subsets of an ensemble of diverse fine-tuned weights from the same initial pre-training with various hyperparameters could be averaged together. 
Since the brute force optimal subset would require a $2^N$ budget of full evaluations on the validation set, the authors proposed a greedy heuristic for finding such a subset with a fixed budget of $N$. 
We can note that model soups have been successfully applied to numerous tasks such as \cite{Chronopoulou2023AdapterSoupWA, MARON2022307}.
Soups properties, out of distribution generalization, adversarial robustnes, etc., have been studied in diverse ways such as \cite{Rame2022ModelRR, ai3040048, Croce2023SeasoningMS, Jaiswal2023InstantSC}. 
Interestingly, soups have been found to be generalizable to mutlitask fine-tuning \cite{Rame2022ModelRR}. 

However, soups' derivative works are limited to the greedy heuristic proposed by original model soups authors \cite{wortsman2022model} and thus limited to a budget $B = N$. 
We propose here to extend this work to any budget $B \in [1, 2^N]$ chosen by the user depending on its resources (low budget for fast soup crafting, higher for more exploration hoping better performances).

Linearizing deep networks using a first-order Taylor expansion around the pretrained weights is a popular strategy employed to simplify computation and getting intuitions on deep networks \cite{liu22, Liu_2023_ICCV, evci2022head2toe}.  
We motivate our algorithm by showing, under this same first-order Taylor expansion, the equivalence between real soups performances and our approximation of soups performances.

\section{Souping on a budget}
\label{sec:souping}

\subsection{Model Soups Global Idea}

Traditionally, to achieve optimal model accuracy, one would (1) train various models using different hyperparameters and (2) select the best-performing model based on a validation set, discarding the others. 
Model soups \cite{wortsman2022model}  revisit the second step of this procedure in the context of fine-tuning large pre-trained models, where fine-tuned models often appear
to lie in a single low error basin.
\cite{wortsman2022model}  found that
averaging the weights of multiple models finetuned with different hyperparameter configurations often improves accuracy and robustness. 
The difference between ensemble and soups is schematized in Figure.~\ref{fig:res}.

\begin{algorithm}[tb]
    \caption{$\mathsf{GreedySoup}$}
    \label{alg:greedy}
 \begin{algorithmic}
    \State {\bfseries Input:} Potential soup ingredients $\{\theta_1, ... , \theta_N \}$ (N models sorted in decreasing order of $\mathsf{ValAcc}\left(\theta_i\right)$).
    \State {\bfseries Procedure:}
    \State $\mathsf{ingredients} \gets \{ \}$
    \State {\bfseries for} $i = 1$ {\bfseries to} $N$ {\bfseries do}
    \State \ \ \ {\bfseries if}  $\mathsf{ValAcc}\left(\mathsf{average}\left(\mathsf{ingredients} \cup \{ \theta_i \}\right)\right)\geq$
    \State \ \ \ \ \ \ \ \ \ \ \ \ \ \ \ \ \ \ $\mathsf{ValAcc}\left(\mathsf{average}\left(\mathsf{ingredients}\right)\right)$ {\bfseries then}
    \State \ \ \ \ \ \ \ $\mathsf{ingredients} \gets \mathsf{ingredients} \cup \{\theta_i \}$
    \State {\bfseries return} $\mathsf{average}\left(\mathsf{ingredients}\right)$
 \end{algorithmic}
\end{algorithm}

Since ensemble of networks finetuned from the same points are likely to but not guaranteed to fall in the same bassine, authors of model soups \cite{wortsman2022model} proposed a greedy procedure to select the adequate subset of models to be averaged and call it : \textit{greedy soup}. 
The greedy soup recipe is summarized in Recipe.~\ref{alg:greedy}.
The greedy soup is constructed by sequentially adding each model as a potential ingredient in the soup, and only keeping the model in the soup if performance on a held out validation set improves. This procedure requires models to be sorted in decreasing order of validation set accuracy, and so the greedy soup can be no worse than the best individual model on the held-out validation set. 
While they suggested that a subset of models used in soups could be averaged in a smooth way and that such a coefficient could be learned in a gradient-based minibatch optimization. They found this procedure to require simultaneously loading all models in memory, which currently hinders its use with large networks and didn't delve deeper in this path to focus on greedy soups. 
We will thus stay in this setup of uniform average of binary selection of models. 
Since \cite{Mandt2017StochasticGD} already shown, under simplifying assumption, that SGD with a fixed learning rate approximately samples from a Gaussian distribution centered at the minimum of the loss, it is perfectly correct to discard the weighted average and focus on uniform average since from a Maximum Likelihood Estimation perspective, the uniform average is the unbiased estimator to recover the center of a gaussian.

\subsection{Formalism}
\label{subsec:formalism}

Let's define $\lbrace p_k \rbrace_{1 \leq k \leq N}$ a set of ponderations such that $\sum_{k = 1}^N p_k = 1$, $\mathcal{L}$ a loss function and $(x_d, y_d)$ our traning set of size $D$ with $x$ an input and $y$ a ground truth label. 
With $w^*_k$ being the $k^{th}$ model weights $w_k$ of $N$ models of $M$ neurons fine-tuned from the same $w_k = w_{init}$, we can define the ensemble loss as : 
\begin{equation}
    \label{eq:ens}
    l_{\ens}(x_d, y_d, w^*_{\ens}) \defeq \mathcal{L}\Big( \sum_{k = 1}^N p_k f(x_d, w^*_k) \Big)
\end{equation}
with $w^*_{\ens}=\{w^*_k\ | \forall k \in [1, K] \}$.

In another hand, we define a soup loss as : 
\begin{equation}
    \label{eq:soup}
    l_{\soup}(x_d, y_d, w^*_{\soup}) \defeq \mathcal{L}\Big( f \big( x_d, \sum_{k = 1}^N p_k w^*_k \big) \Big)
\end{equation}
with  $w^*_{\soup}=\mathsf{soup}(\{p_k\}^*) = \sum_{k = 1}^N p_k w^*_k$.

In our case we are looking for $p_k$ to be either $0$ either uniform $1 / \sum \ind_{p_k > 0}$. 
When all $p_k > 0$ the soup is called uniform soup, since all model weights are now averaged in an uniform fashion. 
Our goal is to find the best soup which could be uniform or only an average of the subset of models selected by $p_k$.

\begin{algorithm}[tb]
    \caption{\RADIN}
    \label{alg:procedure}
 \begin{algorithmic}
    \State {\bfseries Inputs:}  \\~~~~$\mathsf{B}$ (a budget, allowed nb. of full slow evaluation) \\~~$\mathsf{candidates}   \gets \Big[\{p_k\}^1, \ldots, \{p_k\}^c, \ldots, \{p_k\}^C\Big]$ (a list of C random soups candidates defined by their weighting $\{p_k\}^c, \forall c \in [1, C])$).

    \State {\bfseries Procedure:}
\State~$\mathsf{candidates} \gets \mathsf{sorted}(\mathsf{candidates}, \mathsf{key=ApproxValAcc})$
\State~$\mathsf{candidates} \gets \mathsf{topk}(\mathsf{candidates, k=B})$
\State~$\mathsf{candidates} \gets \mathsf{sorted}(\mathsf{candidates}, \mathsf{key=ValAcc})$
    \State {\bfseries return} $\mathsf{soup}(\mathsf{topk}(\mathsf{candidates, k=}1))$
 \end{algorithmic}
\end{algorithm}

\subsection{Methodology}
\label{sub:method}

Here we propose a two stage algorithm for soups crafting. 
From  a fast approximating function we rank soups candidates, then, after selecting the top$k$ (with $k=B$, chosen budget), a slow full inference step will be performed on most promising candidates in order to find the best soup to return, accordingly to its real observed performance. 
Our \textbf{R}essource \textbf{AD}justed soups craft\textbf{IN}g (\RADIN) procedure is summarized in Recipe.~\ref{alg:procedure}. 

To perform the fast candidates performances estimation and ranking step we propose to approximate the logits of averaged models weights by the average of cached models logits. 
In other words, with $\hat{y}_k = f(x, w^*_k)$, we pose : 
\begin{equation}
    \mathcal{L}\Big( f \big( x, \sum_{k = 1}^N p_k w^*_k \big) \Big) \approx \mathcal{L}\Big( \sum_{k = 1}^N p_k \hat{y}_k \Big)
\end{equation}

From our fast estimation function, ones can choose to add any prior as wished to influence the ranking toward any direction. 
Here, we propose, inspired from \cite{Mandt2017StochasticGD} findings, to favorize solutions averaging the maximum of models in the soup. 
We formalize this new objective in the new function : 
\begin{equation}
    \mathcal{L}\Big( \sum_{k = 1}^N p_k \hat{y}_k \Big) + \lambda \sum_{k=1}^N \mathbb{1}_{p_k > 0}
\end{equation}
The $\lambda$ term allows to control the strength of our prior. 
While the top$k$ ranking will favorize larger number of models to be averaged, the real performance evaluation step will have the last word, discarding such a model if the real performance observed is lower than soup with lower number of base models during the second step.

When few models are present in the ensemble, a full exploration of all subsets in a brute force fashion is possible at low cost. 
Such an exploration is not possible as ensemble size grow, since the number of estimates grows exponentially ($2^N$). 
Different methods are possible to find the best subset of models, ranging from simple ones such as Monte-Carlo sampling to more sophisticated ones such as genetic algorithms, reinforcement learning, branch and bound, etc. 
In this paper, we will stay with Monte-Carlo to show the efficiency of our method even with naïve algorithm. 

After generating a list of soup candidate and sorting them accordingly to the previously defined function, the second step of our approach consist in selecting the $B$ mode promising candidates to compute their real score with a full inference over the validation set. 
The best model, accordingly to its real score on validation, is finally chosen and returned as output model soup.

\begin{figure}[!t]
    
  \centering
  \centerline{\includegraphics[width=.87\linewidth]{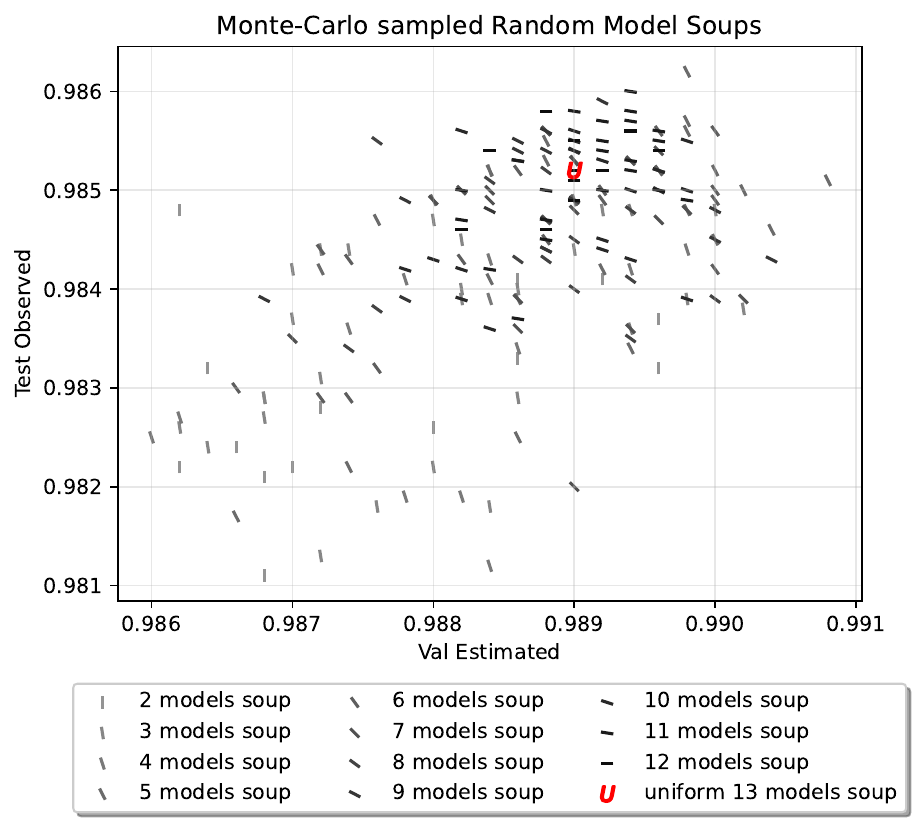}}

  \caption{Correlation between fast estimated performance on validation and real performance observed on test of 200 sampled random soups on Cifar 10. Soups with higher number of models tend to perform generally better than soups with lower number of models. The number of models involved in each soup candidate is indicated by the angle of the marker.}
  \label{fig:monte_carlo_sampled_random_model_soups_cifar10}
\end{figure}

While the Taylor trick allows a fast approximation of model soups using logits ensemble, a two-step approach using slow fine-grained re-ranking at a budget is still necessary since by not being equal but only correlated (Figure.~\ref{fig:monte_carlo_sampled_random_model_soups_cifar10}) the fast approximation may overestimate or underestimate the real performance of a soup.

\subsection{Theoretical Foundations}
\label{subsec:proof}

Using a similar approach as in \cite{evci2022head2toe}, we show that the weighted average of logits of several finetuned models (denoted as the ensembling strategies) is equivalent to the logits of the weighted average of the same finetuned models weights with the exact same ponderatation.

Our algorithm relies on the hypothesis that the Taylor expansion of these two loss functions at $W_{\init} = (w_{\init}, \ldots , w_{\init})$ agree at order 1. 
We propose here a complete proof to validate the intuition behind our algorithm.

Let $N, M \geq 1$ be integer and $\Hom_{\mathbb{R}} (\mathbb{R}^N, \mathbb{R}^M)$ be the space of $\mathbb{R}$-linear maps $\mathbb{R}^N \rightarrow \mathbb{R}^M$. 
If $\varphi : \mathbb{R}^N \rightarrow \mathbb{R}^M$ is a differentiable function at a point $P \in \mathbb{R}^N$, then we let ${d_P \varphi \in \Hom_{\mathbb{R}}(\mathbb{R}^N, \mathbb{R}^M)}$ denote its differential.

The first order Taylor expansions at $W_{\init}$ reads: 
\begin{equation*}
\mathcal{L}_{\ens} (W) \sim \mathcal{L}_{\ens} (W_{\init}) + d_{W_{\init}} \mathcal{L}_{\ens} (W - W_{\init})
\end{equation*}
\begin{equation*}
\mathcal{L}_{\soup} (W) \sim \mathcal{L}_{\soup} (W_{\init}) + d_{W_{\init}} \mathcal{L}_{\soup}  (W - W_{\init}).
\end{equation*}

\noindent\textbf{Proposition :} the first order Taylor expansions at $W_{\init}$ of $\mathcal{L}_{\ens}$ and $\mathcal{L}_{\soup}$ are equal. \\

\noindent\textbf{Proof :} \\
By definition, we have $\sum_{k} p_k = 1$ and thus
\begin{align*}
\mathcal{L}_{\ens} (W_{\init} ) &= \mathcal{L} \Big( \sum_{k=1}^N p_k f(w_{\init})  \Big) = \mathcal{L} \big( f(w_{\init})  \big) \\
&= \mathcal{L} \Big( f \Big( \sum_{k=1}^N p_k w_{\init} \Big)  \Big) = \mathcal{L}_{\soup} (W_{\init} ).
\end{align*}

\noindent From this it follows that, to prove our claim, it suffices to show that
\begin{equation} \label{eq2}
d_{W_{\init}} \mathcal{L}_{\ens} = d_{W_{\init}} \mathcal{L}_{\soup}.
\end{equation}
We define functions $s$ and $F$ by $s (v_1, \ldots , v_N) = \sum_{k} p_k v_k$ and $F : (v_1, \ldots, v_N) \mapsto (f(v_1), \ldots , f(v_N))$. By definition, we have $\mathcal{L}_{\ens} = \mathcal{L} \circ s \circ F$ and $\mathcal{L}_{\soup} = \mathcal{L} \circ f \circ s$. Applying the chain rule, we have both
\begin{equation*}
d_{W_{\init}} \mathcal{L}_{\ens}= d_{W_{\init}} (\mathcal{L} \circ s \circ F) = d_{s \circ F (W_{\init})} \mathcal{L}  \circ  d_{W_{\init}} (s \circ F)
\end{equation*}
\begin{equation*}
d_{W_{\init}} \mathcal{L}_{\soup}= d_{W_{\init}} (\mathcal{L} \circ f \circ s) = d_{f \circ s (W_{\init})} \mathcal{L}  \circ  d_{W_{\init}} (f \circ s)
\end{equation*}

By definition, we have $\sum_{k} p_k = 1$ and this yields
\begin{align*}
s \circ F (W_{\init}) &= \sum_{k = 1}^N p_k f(w_{\init}) = f ( w_{\init}) \\
&= f \Big( \sum_{k = 1}^N p_k w_{\init} \Big) = f \circ s(W_{\init}).
\end{align*}
In particular, we deduce that
\begin{equation*}
d_{s \circ F (W_{\init})} \mathcal{L} = d_{f \circ s (W_{\init})} \mathcal{L}.
\end{equation*}

Now, we can see that, if we prove
\begin{equation} \label{eq1}
d_{W_{\init}} (s \circ F) = d_{W_{\init}} (f \circ s) (W_{\init}),
\end{equation}
then \eqref{eq2} follows and this proves the claim. 

Let $(e_1, \ldots, e_N)$ be the canonical basis of $\mathbb{R}^N$ and let ${1 \leq i \leq N}$ be an integer. On one hand, by chain rule, we have
\begin{align*}
d_{W_{\init}} (s \circ F) (e_i) &= d_{F(W_{\init})} s \circ d_{W_{\init}} F (e_i) \\
&= d_{F(W)} s \big( 0, \ldots, 0, f'(w_{\init}), 0, \ldots , 0 \big) \\
&= p_i f'(w_{\init}).
\end{align*}
On the other hand, by chain rule, we have
\begin{align*}
d_{W_{\init}} (f \circ s) (e_i) &= d_{s(W_{\init})} f \circ d_{W_{\init}} s (e_i) = d_{s(W_{\init})} f (p_i) \\
&= f'  (s(W_{\init})) p_i  = f' \Big( \sum_{k =1}^N p_k w_{\init} \Big) p_i \\
&= p_i f '(w_{\init}).
\end{align*}
This proves that the two linear maps $d_{W_{\init}} (s \circ F)$ and $d_{W_{\init}} (f \circ s)$ are equal on the canonical basis or $\mathbb{R}^N$. Hence they are equal: 
This proves \eqref{eq1} and concludes the proof of the claim. \hfill $\qed$~

\begin{figure}[tb]

  \centering
  \centerline{\includegraphics[width=1.\linewidth]{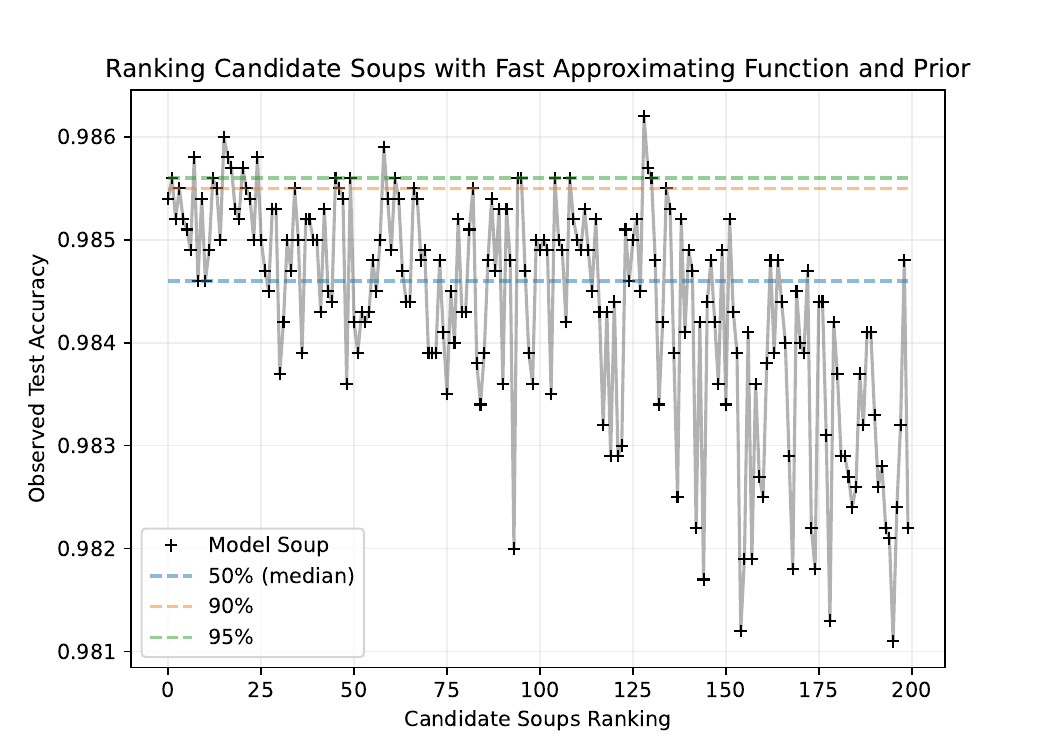}}

  \caption{Soups candidates are ranked by their fast estimated performance. Lower rank indicate most promising candidates. One can observe that the fast approximation ranking allow to filter the poorly performing candidates. In the \RADIN~procedure, only the $B$ first candidates undergo full evaluation, from which the highest-performing soup is selected based on actual performance metrics.}
  \label{fig:candidate_soups_ranking}
\end{figure}

\begin{table}[ht]

\caption{Comparison of performances between uniform soups, greedy procedure and our approach at different budgets $B$. Our soup with prior at budget $B = 1$ correspond to the uniform soup. The oracle correspond to the best test score observed on the test set (cheating, not the validation set) by evaluating up to $B = 200$.} 
\begin{tabular}{@{}l|ll|ll@{}}
\toprule
Budget ($B$)       & \multicolumn{2}{c}{$B = 1$} & \multicolumn{2}{c}{$B = 25$} \\ \midrule
Method        & Cifar 10    & ImageNet    & Cifar 10     & ImageNet    \\ \midrule
Greedy \cite{wortsman2022model} &         98.17 & 74.990 & 98.58 & 78.418 \\ \midrule
Our w/o prior &         98.51 & 77.962 & \underline{\textbf{98.62}} & \underline{\textbf{78.698}} \\
Our w prior   & \underline{\textbf{98.52}} & \underline{\textbf{78.208}} & 98.58 & 78.562 \\ \bottomrule
\textit{Oracle} & \textit{98.62} & \textit{78.726} & \textit{98.62} & \textit{78.726} \\
\end{tabular} 
\label{tab:scores_budget}
\end{table}

\section{Experiments}

\subsection{Experimental Setup}


In this section, we assess the effectiveness of our algorithm by comparing it with the \textit{greedy soups} baseline algorithm suggested by \cite{wortsman2022model}. 
We used the pre-trained CLIP ViT-B/32 transformer with contrastive supervision from image-text pairs as our base model ($w_{init}$). 
For each dataset (Cifar 10, ImageNet), we randomly selected one of an ensemble of 13 fine-tuned end-to-end models provided by \cite{wortsman2022model}, whose hyperparameters information can be found in the original article. 
We utilize the same 2\% of the ImageNet training set as a held-out validation set for building soups as \cite{wortsman2022model} in order to make a fair comparison between the baseline and our technique, since the official ImageNet validation set is typically used as test set.

We consider the uniform average of all the ensemble weights and greedy soups as baseline. 
Since the uniform average is not iterative and the greedy soups have a number of iteration limited, we keep the last value encountered for comparison at higher number of iterations (higher budget). 
For our approach, we compare results obtained with and without using the prior, by setting $\lambda$ respectively to $0$ and $1$.

We sampled 200 candidates without overlap in a Monte-Carlo fashion, with the size of the subset being evenly sampled in the range of $[2, N -1]$. 
For evaluation, we excluded the uniform soup to show the ability of our method to recover a similar performing soup with low budget $B = 1$. 
However, in real world application we recommend adding it to candidates since that with prior $\lambda = 1$ at a budget $B = 1$ this uniform soup would be picked (highering our score on ImageNet while maintaining similar performance on Cifar 10).

\subsection{Results}

\begin{figure}[ht]

  \centering
  \centerline{\includegraphics[width=.8\linewidth]{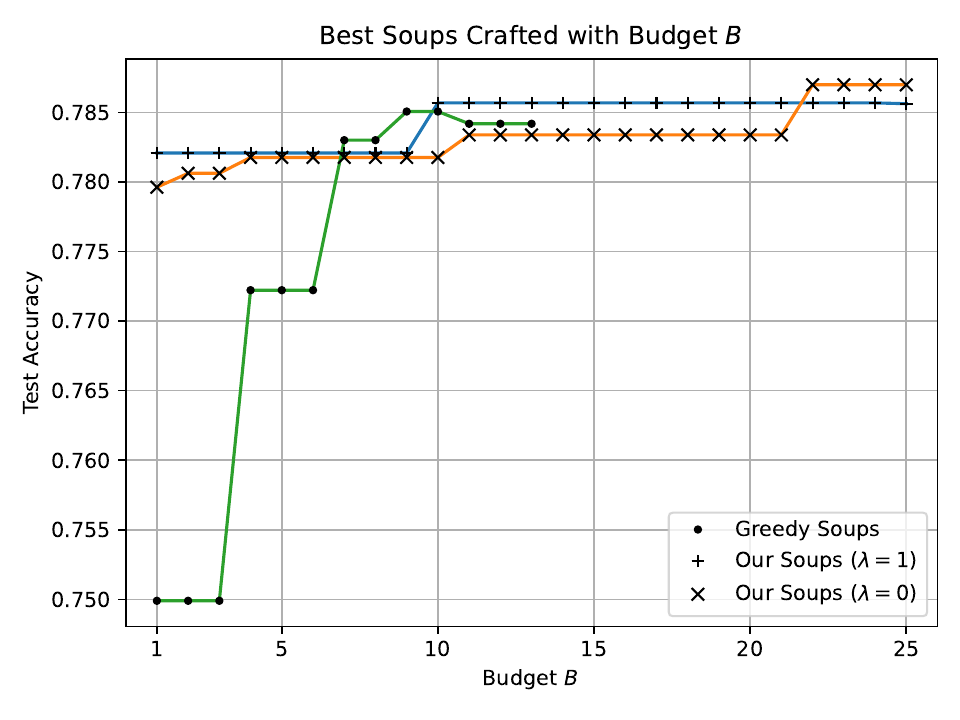}}

  \caption{Performance comparison of model soups on ImageNet at various budget levels $B$. While performances are comparable at higher budgets, \RADIN~outperforms greedy soups at reduced budgets. Notably, introducing the prior $\lambda$ marginally enhances the quality of soups identified at these lower budgets.} 
  \label{fig:imagenet_our_vs_greedy_vs_baseline}
\end{figure}

We report in  Table.~\ref{tab:scores_budget} results on Cifar 10 and ImageNet with baseline and our method at two different budget values ($B = 1, B = 25$). 
To get a finer look at the evolution of performance of different algorithms over iteration, we plot on Figure.~\ref{fig:imagenet_our_vs_greedy_vs_baseline} the evolution of performances for each budget $B \in [1, 25]$ on ImageNet.

By evaluating of the 200 soups candidates on Cifar 10, we can observe on Figure.~\ref{fig:candidate_soups_ranking} that our procedure effectively rank the candidates and filter out the poorly performing ones (bad performance prediction $\approx$ bad real performance).

Interestingly, while we reported a lower correlation between the 200 observed candidates performances on test for Cifar 10 to be lower with the prior than without it (Spearman correlation $0.5625 < 0.6968$ statistically significant with $p$-value $< 0.01$) we observed that using prior tend to perform better at lower budget. 
At higher budget, the removing the prior allow exploration of less conservative candidates by sampling highly different subsets of smaller size (less overlap with smaller subset size). 
This trend was found to be similar for ImageNet as ones can observe in Table.~\ref{tab:scores_budget} and Figure.~\ref{fig:imagenet_our_vs_greedy_vs_baseline}.

\begin{figure}[ht]

  \centering
  \centerline{\includegraphics[width=.8\linewidth]{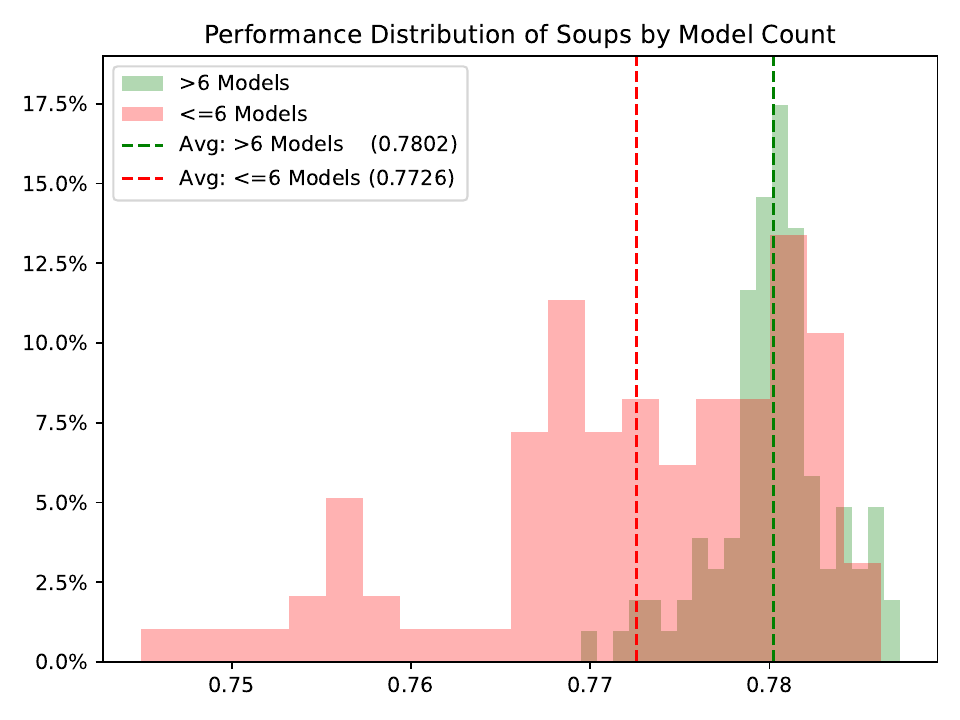}}

  \caption{
  Performances distribution of model soups using high ($> 6$) and low ($\leq 6$) number of models on ImageNet.} 
  \label{fig:perforamnce_distribution_soups_by_model_count_imagenet}
\end{figure}

\subsection{Usefulness of the prior} 

In general results are difficult to discuss to the low difference between every method making each of them non statistically distincitve (better of lower). 
Only the supperior performance of our method at a budget $B = 1$.


\begin{figure}[ht]

  \centering
  \centerline{\includegraphics[width=.8\linewidth]{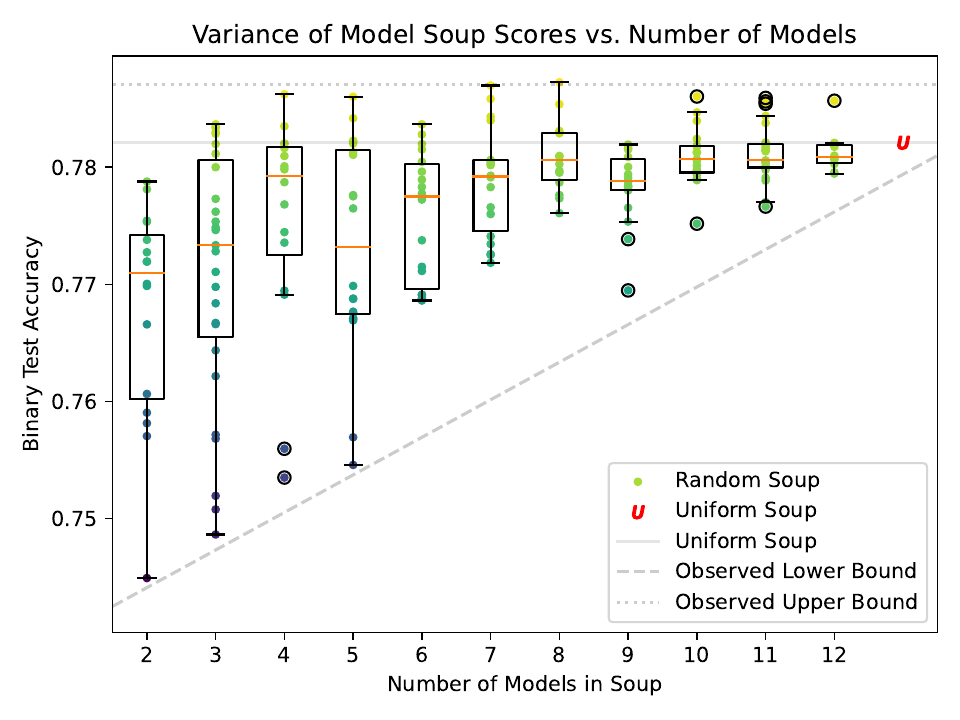}}

  \caption{Visualization of the variance in performances for randomly sampled soup per number of models on ImageNet. } 
  \label{fig:variance_per_count}
\end{figure}

We propose here two visualizations to show the motivation behind the use of the prior.
Firstly, we propose a coarse visualization in   Figure.~\ref{fig:perforamnce_distribution_soups_by_model_count_imagenet} to observe and test stastically the difference between the distribution scores of soups using more than 6 models versus others. 
Due to the non-normal distribuytion of the data, the Mann-Whitney U, a non-parametric test, has been used to compare the difference of performances of soups models using a high and a low number of models in the solution. 
With a $p$-value of $1.034e10-9$ the difference of means are statistically significative.
Secondly, we propose a more fine-grained visualization in Figure.~\ref{fig:variance_per_count} by plotting soups scores variance by number of models used in each soup. 
Such a result is in adequation with \cite{Le2022ModelAI} theoretical work mathematically showing that model averaging is asymptotically better than model selection for prediction. 
However we can note that while the worst performance per count (Observed Lower Bound) is diminishing such as the score variances as the number of models in a soup increase, the best performance do not follow such a trend. 
Indeed, the best soup per count seems to be distributed/plateaued  on a flat line (Observed Upper Bound) with a slightly better performance than the uniform soup. 
The greedy algorithm such as our \RADIN~purpose is to reach such performances. 
In these conditions, 
using the number of models count as prior does not sound useful.

\section{Conclusion}
\label{sec:conclusion}

The ability to efficiently combine models, or craft \textit{model soups}, holds significant potential for performance enhancement. 
This study introduced \RADIN, a novel, resource-adjusted procedure designed to address the complexities inherent in crafting model soups. 
Built upon theoretical foundations, \RADIN~offers an alternative to traditional greedy methodologies, focusing on the potential of ensemble logits for approximating model soup performance. 
Our first attempt is based on naive Monte-Carlo candidates sampling and provided competitive results at low computational budget. 
We think future research may combine our framework with more advanced algorithms to increase performances at higher budgets.


\bibliographystyle{IEEEbib}
\bibliography{strings,refs}

\end{document}